\documentclass{article}

\usepackage[dblblindworkshop, final]{neurips_2025}

\workshoptitle{MATH-AI}

\usepackage[utf8]{inputenc} % allow utf-8 input
\usepackage[T1]{fontenc}    % use 8-bit T1 fonts
\usepackage{hyperref}       % hyperlinks
\usepackage{url}            % simple URL typesetting
\usepackage{booktabs}       % professional-quality tables
\usepackage{amsfonts}       % blackboard math symbols
\usepackage{float}          % for [H] float placement
\usepackage{nicefrac}       % compact symbols for 1/2, etc.
\usepackage{microtype}      % microtypography
\usepackage{xcolor}         % colors
\usepackage{amsmath}
\usepackage{multirow}
\usepackage{tabularx} % in preamble

\usepackage{listings}
\usepackage{amsmath, amssymb} % For \neg, \lor, \land, etc.

\definecolor{codegreen}{rgb}{0,0.6,0}
\definecolor{codegray}{rgb}{0.5,0.5,0.5}
\definecolor{codepurple}{rgb}{0.58,0,0.82}
\definecolor{backcolour}{rgb}{0.95,0.95,0.92}

\lstdefinelanguage{lean}{
  keywords={theorem, by, rw, exact, Prop},
  keywordstyle=\color{blue}\bfseries,
  morecomment=[l]--,
  commentstyle=\color{codegreen},
  stringstyle=\color{codepurple},
  sensitive=true,
  morestring=[b]",
}

\title{Combining Textual and Structural Information for Premise Selection in Lean}

\author{%
  Job Petrovčič \\
  Faculty of Mathematics and Physics, University of Ljubljana\\
  Jadranska 21, 1000 Ljubljana, Slovenia \\
  \texttt{jp10210@student.uni-lj.si} \\
  \And
  David Narváez \\
  Faculty of Mathematics and Physics, University of Ljubljana\\
  Jadranska 21, 1000 Ljubljana, Slovenia \\
  \texttt{david.narvaez@fmf.uni-lj.si} \\
  \And
  Ljupčo Todorovski \\
  Faculty of Mathematics and Physics, University of Ljubljana\\
  Jadranska 21, 1000 Ljubljana, Slovenia \\
  Department of Knowledge Technologies, Jožef Stefan Institute \\
  Institute for Mathematics, Physics and Mechanics \\
  \texttt{ljupco.todorovski@fmf.uni-lj.si} \\
}

\begin{document}

\maketitle

\begin{abstract}
Premise selection is a key bottleneck for scaling theorem proving in large formal libraries. Yet existing language-based methods often treat premises in isolation, ignoring the web of dependencies that connects them. We present a graph-augmented approach that combines dense text embeddings of Lean formalizations with graph neural networks over a heterogeneous dependency graph capturing both state–premise and premise–premise relations. On the LeanDojo Benchmark, our method outperforms the ReProver language-based baseline by over $25\%$ across standard retrieval metrics. These results suggest that relational information is beneficial for premise selection.
% Previous: These results demonstrate the power of relational information for more effective premise selection.
\end{abstract}

\section{Introduction and related work}

Recent advances in artificial intelligence, particularly large language models (LLMs), have demonstrated increasing efficacy in formal mathematics and interactive theorem proving \citep{polu2020generative, jiang2021lisa, xin2024advancing}. A central task in this domain is \emph {premise selection}: retrieving relevant theorems and definitions from extensive libraries to guide the proofs of new theorems. Effective premise selection underpins many automated reasoning tools, such as Sledgehammer \citep{bohme2010sledgehammer} and modern AI-driven provers for Lean \citep{song2023towards}.

LLM-based approaches such as ReProver~\citep{yang2023leandojo} use dual-encoder models to map proof states and premises into a shared vector space, retrieving relevant premises via dense ByT5 embeddings. However, they ignore the structural relationships in formal libraries: the references and dependencies among axioms, theorems, lemmas, and definitions (entries), which offer strong prior knowledge for guiding proofs. Attempts to exploit this structure include~\citep{10.5555/3666122.3668329} and \citep{ferreira-freitas-2020-premise}, but the former neglects Lean~4’s state–tactic framework, while the latter tackles premise selection only in natural text.

We propose a methodology to integrate structural information into language-based premise retrieval for Lean~4. Our contributions are the following:

\begin{itemize}
    \item We extend LeanDojo's dataset extraction to construct a heterogeneous dependency graph from Lean source files that allows for graph-enhanced premise selection in Lean's tactic mode.
    \item We design a simple relational graph neural network (RGCN) architecture to propagate graph structural information, producing graph-aware premise and proof-state representations. We demonstrate improved performance over the ReProver baseline on the LeanDojo benchmark, highlighting the benefit of incorporating dependency structure.
\end{itemize}

\section{State-premise dependency graph for Lean}

\subsection{Augmented data extraction from the Mathlib Library}
In tactic-based interactive theorem proving, the \emph{proof state} at any point inside a proof represents the statement that currently remains to be proven (the \emph{goal}), together with its local hypotheses. In Lean~4 one enters the tactic environment with the \texttt{by} keyword. When a user applies a tactic, this state is updated to reflect the resulting subgoals. Figures~\ref{lst:state1} and \ref{lst:state2} show the proof states as displayed by Lean~4 for the example theorem in Figure~\ref{fig:premise_example}. The first proof state in Figure~\ref{lst:state1} is the initial state after the \texttt{by} keyword. After applying the tactic \lstinline[language=lean]|rw [$\leftarrow$ not_or]| to this state, we obtain the second proof state depicted in Figure~\ref{lst:state2}.

\begin{figure}[h]
\centering
\begin{lstlisting}[language=lean, caption={Lean theorem using listings}]
theorem premise_example -- (a) premise node
  (p q : Prop) (h : $\neg$ (p $\lor$ q)) -- (b) signature hypotheses
  : $\neg$ p $\land$ $\neg$ q := by -- (c) signature goal
  rw [$\leftarrow$ not_or] -- (d) proof-step dependency
  exact h -- (d)
\end{lstlisting}

\caption{A hypothetical Lean~4 theorem illustrating the extraction of the graph components. \textup{(a)}~The theorem \texttt{premise\_example} becomes a premise node. \textup{(b)}~Signature hypotheses (with edges to premises \texttt{Or, Not}) and \textup{(c)}~the goal (with edges to premises \texttt{And, Not}) define signature dependency edges. \textup{(d)}~The tactic application creates a proof-dependency edge to premise \texttt{not\_or}. The proof states (e.g., after the \texttt{by} keyword, Figure~\ref{lst:state1}) become graph nodes linked to premises in their local hypotheses and goals.}
\label{fig:premise_example}
\end{figure}

\begin{figure}[h]
    \centering
    % First figure
    \begin{minipage}[t]{0.45\textwidth} % [t] aligns at the top
        \begin{lstlisting}
p q : Prop -- (e)
h : $\neg$ (p $\lor$ q)  -- (e)
$\vdash$ $\neg$ p $\land$ $\neg$ q  -- (f)
\end{lstlisting}

        \caption{Initial proof state immediately after \texttt{by}. \textit{State nodes} are created for proof states. The initial state after \texttt{by} (in Figure~\ref{lst:state1}) is linked to its local hypotheses (\texttt{[Or, Not]}) and goal premises (\texttt{[And, Not]}).} 
        \label{lst:state1}
    \end{minipage}
    \hfill
    % Second figure
    \begin{minipage}[t]{0.45\textwidth} % [t] aligns at the top
        \begin{lstlisting}
p q : Prop -- (g)
h : $\neg$ (p $\lor$ q) -- (g)
$\vdash$ $\neg$ (p $\lor$ q) -- (h)
\end{lstlisting}
        \caption{The updated proof state obtained after applying the tactic \protect\lstinline[language=lean]|rw [$\leftarrow$ not_or].|}
        \label{lst:state2}
    \end{minipage}
\end{figure}

\textbf{Base dataset: proof states and premises}  
In the LeanDojo machine learning framework~\citep{yang2023leandojo}, premise selection is formulated as the following task. For each proof state $s$, given its textual representation $x_s$, we aim to identify the list of relevant premises $y_s$ from the (Mathlib \citep{mathlib}) library that will be used in the next tactic. For example, for the proof state in Figure~\ref{lst:state1}, the tactic uses the premise \texttt{not\_or}. Although only one premise appears in this example, in general, a tactic application may use multiple premises or none. Let $S$ denote the set of proof states, and $X_S$ the text representations for each $s \in S$.
  
Let $P$ be the set of all potential premises available in the library. Each premise $p \in P$ has a library definition in Lean code, and we denote by $X_P$ the code (text) representations of the premises in $P$. For example, in Figure~\ref{fig:premise_example}, the entire block defines the premise \texttt{premise\_example}. The dataset in~\citep{yang2023leandojo} can therefore be summarized as the tuple $(X_P, X_S, Y_S)$, where $Y_S$ denotes the set of all lists of relevant premises $y_s$.

\textbf{Graph structure} Definitions of premises, as well as proof states, may reference previously defined premises. The graphical user interface for Lean~4 allows a user to navigate to a previously defined premise by clicking on its symbol in the formalization code of a premise or a proof state. We treat the underlying navigation links as references in our graph.

By merging $V = P \cup S$ as the node set, $X = X_S \cup X_P$ as the textual representations of the corresponding Lean formalizations, and references $E \subseteq V \times \mathcal{R} \times V$ as the edge set, we can summarize a Lean library as a text-attributed \emph{directed} graph $G = (V, X, E)$. Here $\mathcal{R}$ is the set of relation types described in the next paragraph. We extend LeanDojo’s open-source extraction framework \citep{yang2023leandojo} to query the proof assistant for this additional information and extract this full dataset, i.e., the tuple $(G, Y_S)$. The dataset (for Mathlib) thus now includes $(X_P, X_S, Y_S)$ from \citep{yang2023leandojo}, but adds the additional premise-premise and premise-state edges, forming a directed graph. The modified extraction code and the learning pipeline used is available at \url{https://github.com/JobPetrovcic/GNNReProver/tree/lighweight}.

\textbf{Relation types}  
References can be categorized into different types depending on where premises appear in a definition or proof state. Each definition consists of: 
\begin{enumerate}
    \item The name (\textup{(a)} in the example in Figure~\ref{fig:premise_example}).
    \item The signature---which itself splits into local hypotheses \textup{(b)} and the goal \textup{(c)}. 
    \item \label{par:relations:proofs} The proof (\textup{d}), required for theorems, lemmas, and definitions, but not for axioms.
\end{enumerate}
A relation is thus assigned a type based on the positions of its occurrence, and we denote the set of these relation types by $\mathcal{R}$.

Proof states follow a similar structure: they consist of local hypotheses (\textup{(e)} and \textup{(g)} in the examples in Figure~\ref{lst:state1} and \ref{lst:state2}) and goals (\textup{(f)} and \textup{(h)}). The proof component is represented by the next tactic applied to this state. The premises the tactic uses are the lists $y_s$ introduced earlier.

\textbf{File and import graph} To remain consistent with the framework of~\citep{yang2023leandojo}, we also use the separate directed graph of imports between the files and a map between entries, states, and the files where they were defined. We do not use this information during training, and it is not employed by the model. We leave the utilization of this information for future work. However, during evaluation, this graph is used to restrict the premise selection only to premises that would have been available to the model at that point in the file: premises from \textit{imports}, and premises defined \textit{before} the current entry. For complete details, see~\citep{yang2023leandojo}. 

\subsection{Dataset statistics}
Table~\ref{tab:graph_stats} in the appendix summarizes the graph statistics for the augmented LeanDojo benchmark extracted from the Mathlib library commit \texttt{29dcec074de168ac2bf835a77ef68bbe069194c5}, the one used in the official repository of~\citep{yang2023leandojo}. This allows us to directly compare our results with the results of their approach.

\section{Methodology}

\subsection{GNN-augmented premise retrieval}
We can now formulate premise selection as learning a scoring function $f \colon S \times P \to \mathbb{R}$ that measures the relevance of premise $p \in P$ for proof state $s \in S$. To this end, we apply GNN-refined embeddings as follows.

\textbf{Step 1: initial text embeddings.} The textual representations of premises $x_p$ and states $x_s$ are initially embedded using ReProver's ByT5 dual-encoder \citep{yang2023leandojo}, yielding initial node feature vectors $\mathbf{h}_p^{(0)}$ and $\mathbf{h}_s^{(0)}$, respectively. 

\textbf{Step 2: GNN-based refinement.} We employ a \textit{Relational Graph Convolutional Network} (RGCN) \citep{schlichtkrull2018rgcn} to propagate information over the heterogeneous directed graph. Each premise node's embedding $\mathbf{h}_p$ is updated iteratively using
$$
    \mathbf{h}_p^{(l+1)} = \sigma \Bigg( \mathbf{W}_0^{(l)} \mathbf{h}_p^{(l)} + \sum_{(u, r, p) \in E} \frac{1}{\mathcal{N}_r(p)} \mathbf{W}_r^{(l)} \mathbf{h}_u^{(l)} \Bigg)
$$
over $L$ layers, where $\mathcal{N}_r(p) = |\{u | (u, r, p) \in E, u \in P, p \in P\}|$, $\sigma$ is an activation function, and $\mathbf{W}_{r}^{(l)}$ are trainable RGCN weight matrices.

\textbf{Step 3: GNN-refined state encoding.} At retrieval, the proof state is treated as a temporary query node $s$ connected to its premises. Using the same architecture with different weights, the same one-step message passing produces the embeddings of the states $\mathbf{h}_s^{(l+1)'}$ by aggregating embeddings $\{\mathbf{h}_p^{(l)'}\}$ of premises referenced by the state. Note that the edge directions prevent information flow from states to dependencies or future premises to prior dependencies.

\subsection{Training objective}
We use the InfoNCE \citep{infonce, rusak2024infonce} loss, which for a given list $y_s$ of valid premises, calculates as:
$$
\mathcal{L}_{\mathrm{InfoNCE}} = - \frac{1}{\sum_{s \in S} |y_s|} \sum_{s\in S} \sum_{p\in y_s}  \ln \frac{e^{\mathrm{sim}(\mathbf{h}_s^{(L)'},\mathbf{h}_p^{(L)})/\tau}}{\sum_{k\in P} e^{\mathrm{sim}(\mathbf{h}_s^{(L)'},\mathbf{h}_k^{(L)})/\tau}}, 
\quad 
\mathrm{sim}(\mathbf{u},\mathbf{v}) = \frac{\mathbf{u}\cdot \mathbf{v}}{\|\mathbf{u}\|\|\mathbf{v}\|},
$$
where the hyperparameter $\tau$ is a scalar temperature. Note that by minimizing this loss, we train a model of the scoring function $f(s, p) = \mathrm{sim}(\mathbf{h}_s^{(L)'}, \mathbf{h}_p^{(L)})$. The InfoNCE loss contrasts positive and negative premises for each state, summing probabilities over multiple valid premises. The GNN is trained transductively on the full premise graph, excluding all edges from proof relations (the \hyperref[par:relations:proofs]{third item} in the enumeration of relation types in Section~2.1) to prevent trivial memorization and promote learning of general structural patterns.

\subsection{Other optimizations}
We use an ensemble of six independently-trained models (N=6), averaging their outputs to form the final prediction. This approach mitigates initialization sensitivity, which is a known issue in GNNs~\citep{gnn_init}. In addition, we apply exponential model averaging (EMA) to further improve performance and generalization~\citep{ema}. Finally, compared to LLMs, GNNs use relatively less memory, removing the need to sample negative examples. We thus simply use all other premises in the library as negatives. 

\section{Experiments and results}

\subsection{Experimental setup}
For our experiments, we use the LeanDojo Mathlib benchmark dataset with graph augmentation. We adopt the same train/validation/test split as~\citep{yang2023leandojo} (the “random” split) and evaluate with the same metrics: Recall@1, Recall@10, and Mean Reciprocal Rank (MRR). Our baseline is the ReProver retriever with a ByT5-small encoder. We tune our model’s hyperparameters using Optuna on the validation set (see Appendix~\ref{appendix:hyperparameters}). The number of GNN layers used is $L=2$.

\subsection{Results}
Results reported in Table~\ref{tab:results} show that our GNN-augmented retriever outperforms the baseline across all metrics. Note that the baseline results differ from those reported in \citep{yang2023leandojo}. On GitHub\footnote{\url{https://github.com/lean-dojo/ReProver/discussions/51}}, the authors explain that this is potentially due to improvements made to the ReProver system from the time of their publication.

\begin{table}[h]
\centering
\begin{tabularx}{\textwidth}{X c c c c c c}
\toprule
Model & R@1 & $\Delta$ & R@10 & $\Delta$ & MRR & $\Delta$ \\

\midrule
ReProver (Baseline) & 13.42\% & -- & 39.60\% & -- & 0.3283 & -- \\
GNN-augmented retriever (Ours) & 17.98\% & +33.98\% & 50.04\% & +26.36\% & 0.4095 & +24.73\% \\
\textbf{Ours + EMA} & \textbf{18.31\%} & \textbf{+36.44\%} & \textbf{50.33\%} & \textbf{+27.10\%} & \textbf{0.4140} & \textbf{+26.10\%} \\
\bottomrule
\end{tabularx}
\caption{Retrieval performance on LeanDojo test set. Relative improvements $\Delta$ are w.r.t.~ReProver (Baseline).\label{tab:results}}
\end{table}

\subsection{Ablations}
To understand the contribution of different components in our model, we perform an ablation study on the augmented LeanDojo data set. Specifically, we examine the effect of excluding the graph between contexts and premises and the graph between premises and premises. Table~\ref{tab:ablation} summarizes the results of the ablations. Note that due to computational cost, no ensembling was used in all cases for fair comparison and thus the results for No ablation in Table~\ref{tab:ablation} do not match those in Table~\ref{tab:results}.

\begin{table}[H]
\centering

\begin{tabular}{lccc}
\toprule
\textbf{Model} & \textbf{R@1} & \textbf{R@10} & \textbf{MRR} \\
\midrule
No ablation & 17.43\% & 48.52\% & 0.4010 \\
No ablation + EMA & 17.74\% & 48.70\% & 0.4048 \\
\midrule
Context graph ablation & 17.30\% & 49.99\% & 0.4008 \\
Context graph ablation + EMA & 17.58\% & 50.13\% & 0.4057 \\
\midrule
Premise graph ablation & 17.45\% & 49.30\% & 0.4008 \\
Premise graph ablation + EMA & 18.18\% & 49.98\% & 0.4096 \\
\bottomrule
\end{tabular}
\smallskip
\caption{Ablation study on the LeanDojo test set.\label{tab:ablation}
}
\end{table}

After performing ablations, we realized that removing parts of the dependency graph even improves performance. This leads to the possibility that the utilization of the graph structure is not the main factor behind the superior performance, but rather that this is caused by the different choices of the loss function and sampling strategy. We leave this scrutinization for future work.

\section{Conclusion}

We introduced a graph-augmented language approach to premise selection in Lean. By extracting fine-grained syntactic dependencies and propagating structural information via a GNN, our method produces embeddings that outperform text-based baselines. 

As the ablation section suggests, however, the superior performance might be attributed to the different training paradigm choices rather than the model’s utilization of graph structure. Besides further analysis of the root cause of the improvement, future work includes exploring advanced GNN architectures, such as graph attention networks, to better incorporate structural information. Finally, the model will be evaluated on more realistic splits, such as the LeanDojo "novel" split or a split based on the creation time of premises.

\section{Acknowledgments}

The authors acknowledge the financial support of the Slovenian Research Agency via the Gravity project \textsl{AI for Science}, GC-0001 and via the research core funding No.~P2-0103, the Air Force Office of Scientific Research under award number FA9550-21-1-0024, and the Renaissance Philanthropy via the project \textsl{Bridging AI, Proof Assistants, and Mathematical Data (BRIDGE)}. We are especially grateful to Boshko Koloski for the beneficial discussion related to the training of our model.

\bibliography{references}
\bibliographystyle{plainnat}

\medskip

%{
%\small
%
%
%[1] Alexander, J.A.\ \& Mozer, M.C.\ (1995) Template-based algorithms for
%connectionist rule extraction. In G.\ Tesauro, D.S.\ Touretzky and T.K.\ Leen
%(eds.), {\it Advances in Neural Information Processing Systems 7},
%pp.\ 609--616. Cambridge, MA: MIT Press.
%
%
%[2] Bower, J.M.\ \& Beeman, D.\ (1995) {\it The Book of GENESIS: Exploring
%  Realistic Neural Models with the GEneral NEural SImulation System.}  New York:
%TELOS/Springer--Verlag.
%
%
%[3] Hasselmo, M.E., Schnell, E.\ \& Barkai, E.\ (1995) Dynamics of learning and
%recall at excitatory recurrent synapses and cholinergic modulation in rat
%hippocampal region CA3. {\it Journal of Neuroscience} {\bf 15}(7):5249-5262.
%}

%%%%%%%%%%%%%%%%%%%%%%%%%%%%%%%%%%%%%%%%%%%%%%%%%%%%%%%%%%%%

\appendix

\appendix

\section{Dataset statistic for Mathlib~4}

\begin{table}[H]
\caption{Summary statistics of the augmented LeanDojo Mathlib~4 dependency graph, including node counts and edge types used for premise and state relations.}
\label{tab:graph_stats}
\centering
\begin{tabular}{lll}
\toprule
\textbf{Node Statistic} & \textbf{Value} &  \\
\midrule
Total Number of Nodes & 440,487 & \\
\quad Premise Nodes & 180,907 & \\
\quad State Nodes & 259,580 & \\
\midrule
\textbf{Edge Statistics} & \textbf{Premise-to-Premise} & \textbf{Premise-to-State} \\
\midrule

Signature local hypotheses edges & 652,484 (36.9\%) & 2,670,304 (63.4\%) \\
Signature goal edges & 626,318 (35.4\%) & 1,539,899 (36.6\%) \\
Proof-dependency edges & 490,356 (27.7\%) & / \\
Next tactic premise labels ($Y_S$) & / & 379861 \\
\midrule
Total Edges & 1,769,158 & 4,210,203 \\
\bottomrule
\end{tabular}
\end{table}

\section{Hyperparameters}
\label{appendix:hyperparameters}

\subsection{Hyperparameter tuning}
The hyperparameters for our GNN retrieval model were determined through a two-stage search using the Optuna framework \citep{optuna}. The first stage focused on architectural choices, such as GNN layer type, hidden dimensions, learning rate, and model structure (e.g., separate vs. shared GNNs for premises and contexts). The objective was to maximize the Recall@10 metric on the \emph{training set for the first batch}.

As the best model was prone to overfitting, we introduced a second stage, fixing the best architecture from the previous stage, and performed a fine-grained search for optimal regularization parameters. This included tuning the node feature dropout rate, the edge dropout rate, and the L2 weight decay for the optimizer. The objective in this case was to maximize the Recall@10 metric on the \emph{validation} set.

\subsection{Model and training configuration}
Table~\ref{tab:gnn_config} lists the final configuration of our best-performing model, obtained through a two-stage Optuna search described in the previous section. We train for 120 epochs and select the model with the best validation Recall@10 for evaluation on the test set.

\begin{table}[h!]
\centering
\small
\caption{Final model and training configuration for the GNN-augmented retrieval model.}
\begin{tabular}{llr}
\toprule
\textbf{Category} & \textbf{Parameter} & \textbf{Value / Notes} \\
\midrule
\multirow{6}{*}{GNN} 
 & number of layers & 2 \\
 & hidden size & 1024 \\
 & activation & ReLU \\
 & dropout & 0.256 \\
 & edge dropout & 0.142 \\
 & residual connections & Used \\
\midrule
Loss & InfoNCE temperature & 0.0138 \\
\midrule
\multirow{2}{*}{Optimizer} 
 & learning rate & 0.00499 \\
 & weight decay & 2.359e-5 \\
\midrule
\multirow{2}{*}{Training} 
 & batch size & 1024 \\
 & gradient accumulation & 2 batches \\
 & epochs & 120 \\
\bottomrule
\end{tabular}
\label{tab:gnn_config}
\end{table}

\section{Resources}

We ran all experiments on a cluster with three NVIDIA A6000 GPUs (48 GB each), two Intel Xeon Silver 4410Y CPUs (24 cores, 48 threads), and 512\,GB of RAM.

The initial hyperparameter tuning stage, with reduced training epochs, took about one day, and the subsequent stage about. 12~hours. Final training of the ensemble of six independently trained models took roughly one day, with exponential moving average (EMA) optimization adding negligible time.

%\section{Technical Appendices and Supplementary Material}
%Technical appendices with additional results, figures, graphs and proofs may be submitted with the paper submission before the full submission deadline (see above), or as a separate PDF in the ZIP file below before the supplementary material deadline. There is no page limit for the technical appendices.

%%%%%%%%%%%%%%%%%%%%%%%%%%%%%%%%%%%%%%%%%%%%%%%%%%%%%%%%%%%%

\end{document}